\documentclass{article}

\usepackage[T1]{fontenc}
\usepackage{times}
\usepackage{hyperref}
\usepackage{url}
\usepackage{color}  
\hypersetup{urlcolor=black,linkcolor=black,citecolor=black,colorlinks=true} 
\usepackage{amsmath}
\usepackage{amsfonts}
\usepackage{amssymb}
\usepackage{mathabx}
\usepackage{dcolumn}
\usepackage{graphics}
\DeclareGraphicsExtensions{.png,.jpg}
\usepackage{graphicx}
\usepackage{verbatim}
\usepackage{afterpage}
\usepackage{multirow}
\usepackage{subcaption}
\usepackage{stfloats}
\usepackage[utf8]{inputenc}
\usepackage{algorithm}
\usepackage{algorithmic}
\usepackage{booktabs}
\usepackage{longtable}
\usepackage{multirow}
\usepackage[normalem]{ulem}
\usepackage{setspace,lipsum}

\newcommand{\surl}[1]{{\small\url{#1}}}

\def\-{\discretionary{}{}{}} 

\newcolumntype{d}[1]{D{.}{.}{#1}}

\begin{document}

\title{Semantically-informed distance and similarity measures for paraphrase plagiarism identification\thanks{Preprint of~\cite{thispaper}. The final publication is available at IOS Press through \url{https://content.iospress.com/articles/journal-of-intelligent-and-fuzzy-systems/ifs169483}}}



\author{Miguel A. {\'A}lvarez-Carmona$^1$, Marc Franco-Salvador$^2$,\\ Esa\'{u} Villatoro-Tello$^3$ , Manuel Montes-y-G\'{o}mez$^1$,\\ Paolo Rosso$^4$, Luis Villase\~{n}or-Pineda$^1$\\ 
$^1$\emph{Computer Science Department}\\ \emph{Instituto Nacional de Astrof\'{i}sica, \'{O}ptica y Electr\'{o}nica} \\ \emph{Puebla, M\'{e}xico}\\
$^2$ \emph{Symanto Research, Nuremberg, Germany}\\ 
$^3$ \emph{Information Technologies Department, Universidad Aut\'{o}noma Metropolitana}\\ \emph{Unidad Cuajimalpa, Ciudad de M\'{e}xico, M\'{e}xico}\\ 
$^4$\emph{PRHLT Research Center}\\ \emph{Universitat Polit\`{e}cnica de Val\`{e}ncia, Spain}}

\maketitle
\vspace{-7mm}
\begin{abstract}
Paraphrase plagiarism identification represents a very complex task given that plagiarized texts are intentionally modified through several rewording techniques. Accordingly, this paper introduces two new measures for evaluating the relatedness of two given texts: a semantically-informed similarity measure and a semantically-informed edit distance. Both measures are able to extract semantic information from either an external resource or a distributed representation of words, resulting in informative features for training a supervised classifier for detecting paraphrase plagiarism. Obtained results indicate that the proposed metrics are consistently good in detecting different types of paraphrase plagiarism. In addition, results are very competitive against state-of-the art methods having the advantage of representing a much more simple but equally effective solution.
\end{abstract}



\section{Introduction}
\label{sec:Introduction}
Text plagiarism means including other person's text as your own without proper citation \cite{pandey2015menace}. Nowadays, because of the Web and text editing tools, it is very easy to find and re-use any kind of information \cite{abdi2015pdlk}, causing the plagiarism practice to dramatically increase.

Traditional methods for plagiarism detection consider measuring the word overlap between two texts \cite{lukashenko2007computer}. Using measures such as the Jaccard and cosine coefficients \cite{gomaa2013survey} resulted in a simple but effective approach for determining the similarity between the suspicious and the source texts \cite{hoad2003methods,zechner2009external}.

Likewise, measuring the similarity of texts by means of an edit-distance \cite{levenshtein1966binary,stamatatos2011plagiarism,chatterjee2015edit} or the Longest Common Subsequence (LCS) \cite{gomaa2013survey} resulted in effective approaches. In general, these approaches are very accurate on detecting verbatim cases of plagiarism (i.e., copy-paste), but they are useless to detect complex cases of plagiarism, such as \textit{paraphrase plagiarism}, where texts show significant differences in wording and phrasing.

Detecting paraphrase plagiarism represents a challenging task for current methods since they are not able to measure the \textit{semantic} overlap. Accordingly, some research works have tried to overcome this limitation by proposing the use of knowledge resources such as WordNet \cite{miller1995wordnet} for evaluating the semantic proximity of texts \cite{Biggins2012,Courtney2005,palkovskii2011}. Although these methods have been widely applied for measuring the degree of paraphrases between two given texts, just \cite{palkovskii2011} evaluates its relevance for plagiarism detection. More recently, \cite{brlek2016plagiarism,kim2016bridging} discussed the use of semantic information without depending on any external knowledge resource. Particularly, they proposed using distributive representations, such as word2vec \cite{mikolov2013efficient}, in the task of plagiarism detection. The main drawback of these approaches is that they often need large training sets in order to learn accurate models.

This paper focuses on the detection of paraphrase plagiarism. It proposes two new measures for evaluating the relatedness of two given texts: a semantically informed  similarity measure and a semantically informed edit distance. Both measures can extract the semantic information from WordNet and word2vec. On the top of these measures we trained a classifier for detecting paraphrase plagiarism. In short, the goal of this paper is threefold: \textit{i}) to evaluate the effectiveness of the proposed measures, when using WordNet and word2vec, in the paraphrase plagiarism identification task; \textit{ii}) to investigate the complementarity of both kind of measures for solving the posed task; and \textit{iii}) to determine the effectiveness of the semantically informed measures on detecting specific types of (plagiarism) paraphrases.

The remainder of this paper is organized as follows. Section \ref{sec:Propose} describes the proposed semantically informed measures; Section \ref{sec:setup} describes the used datasets and the experimental setup; Section \ref{sec:Experiments} presents and discusses the obtained results. Finally, Section \ref{sec:Conclusions} depicts our conclusions and some future work directions.

\section{Proposed semantically-informed measures}
\label{sec:Propose}

This section describes the two proposed measures for paraphrase plagiarism identification. Section~\ref{subsec:SemanticallyInformedJaccard} presents a modification of the Jaccard coefficient considering semantic information, whereas Section~\ref{subsec:SemanticallyInformedLevenshtein} describes our semantically informed version of the Levenshtein edit distance.

\begin{figure*}[!t]
  \centering
    \includegraphics[width=350px]{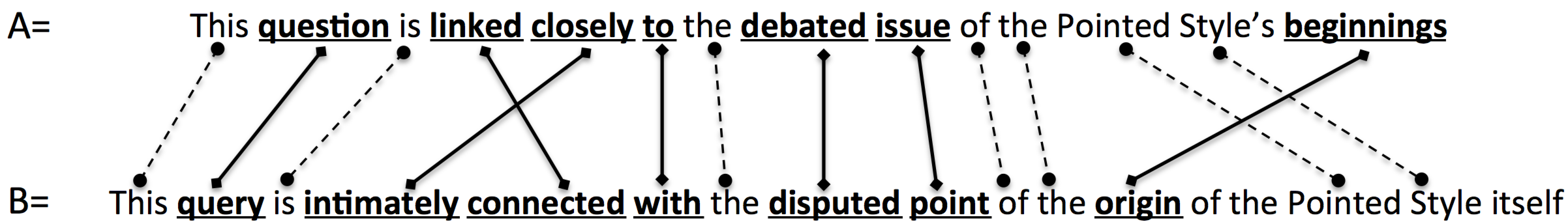}
    \caption{Example of two ($A$ and $B$) semantically related sentences. Dotted lines connect exact matching words whilst solid lines connect semantically related words.}
  \label{fig:SampleSemanticSim}
\end{figure*}

In order to illustrate the limitations of traditional measures and to motivated our proposed modifications, please consider the two sentences from Figure \ref{fig:SampleSemanticSim}. Applying the traditional Jaccard measure it will result in a low similarity, $\text{J}(A,B) = 0.31$, since only 7 terms out of a total of 22 match exactly. Similarly, the classic Levenshtein edit distance will indicate that the sentences are very distant, $\text{ED}(A,B) = 0.70$. Nevertheless, it is evident that these two texts are more similar than these results indicate; they contain several additional pair of terms (solid line connections) that are semantically related but not considered. Therefore, our proposal is to semantically enrich these measures by means of including the similarity degree of non-overlapped words.

\subsection{Semantically-informed similarity measure}
\label{subsec:SemanticallyInformedJaccard}

Let's assume $A$ and $B$ are two documents with vocabularies $V_A$ and $V_B$, and that $V'_A$ and $V'_B$ indicate their non-overlapping words respectively. Their semantic similarity, based on the Jaccard coefficient, is computed as shown in Formula \ref{eq:semanticJaccard}.

\begin{equation}
\label{eq:semanticJaccard}
\text{SJ}(A, B) = \frac{| V_A \cap V_B | + \text{softmatch}(V'_A, V'_B)}{|V_A \cup V_B| - \text{softmatch}(V'_A, V'_B)}
\end{equation}

The function $\text{softmatch}(X, Y)$ accounts for the maximum similarity values between words contained in the sets $X$ and $Y$. For its computation we first measure the similarity $\text{sim}(x,y)$ among all words $x \in X$ and $y \in Y$; this similarity could be measured using WordNet or word2vec. Then, we eliminate irrelevant relations, that is, we set $\text{sim}(x, y) = 0$ if it is not the greatest similarity score for both $x$ and $y$ with any other term. Finally, we accumulate the resulting similarities as indicate by Formula \ref{eq:semanticInfo}.

\begin{equation}
\label{eq:semanticInfo}
\text{softmatch}(X, Y) = \sum_{\forall x \in X, \forall y \in Y} \text{sim}(x,y)
\end{equation}

Continuing with the example from Figure~\ref{fig:SampleSemanticSim}, $V_A'=$ \{question, linked, closely, to, debated, issue, beginnings\} and $V_B'=$ \{query, intimately, connected, with, disputed, point, origin\}. Using WordNet as semantic resource for computing word similarities as described in Section \ref{sec:wordsimilarity}, $\text{softmatch}(A',B')= 6.75$, resulting in $\text{SJ}(A, B) = 0.90$, which in turn reflects a more realistic similarity than the initial estimated value. 

\subsection{Semantically-informed edit distance}
\label{subsec:SemanticallyInformedLevenshtein}

This new measure is based on the Levenshtein edit distance. It also computes the minimum number of operations permitted (generally substitution [S], deletion [D] and insertion [I]) for transforming text $A$ to text $B$. However, different to the traditional version where each operation has unitary cost, our proposal accounts for the semantic similarity between substituted words and determines the impact of inserted/deleted words in the text. The proposed semantically-informed edit distance between two texts $A$ and $B$, of lengths $|A|$ and $|B|$ respectively, is given by $\text{SED}_{A,B}(|A|,|B|)$ where:

\begin{equation}
\small
\label{eq:SemanticLev}
\text{SED}_{A, B}(i,j) = \text{min} \left \{ \begin{matrix} \text{SED}(i-1, j) + \text{dist}(\tau, A_i) & \mbox{D}
\\ \text{SED}(i, j-1) + \text{dist}(\tau, B_j) & \mbox{I}
\\ \text{SED}(i-1, j-1) + \text{dist}(A_i, B_j) & \mbox{S} \end{matrix}\right.
\end{equation}

In this approach the substitution of a word $x$ by a word $y$ has a cost proportional to their semantic distance $\text{dist}(x,y)$. This distance could be measured using WordNet or word2vec as described in Section \ref{sec:wordsimilarity}. Similarly, the insertion or deletion of a word $x$ has a variable cost, which is defined in function of its semantic distance to a predefined general word $\tau$. The idea is that the greater $\text{dist}(\tau,x)$, the more rare is the word $x$, and the more important its contribution of the meaning of the text. 

Following with the example above, the new edit distance between texts $A$ and $B$ is small, $\text{SED}(A,B) = 0.20$, because all words in bold face are substituted by semantically related words, for instance, ``question" by ``query" and ``beginnings" by ``origin". In addition, all removed words, such as ``of", ``the" and ``itself" are very general and, therefore, their deletion do not have a considerable impact.   

\section{Experimental Setup}
\label{sec:setup}

The proposed distance and similarity measures are especially suited to the task of paraphrase plagiarism identification. Accordingly, this section presents the datasets used for their evaluation as well as a description of their configuration for the task. 

\subsection{Datasets.} 
\label{subsec:dataset}
We used the P4PIN corpus\footnote{Available at: \url{http://ccc.inaoep.mx/~mmontesg/resources/corpusP4PIN.zip}} 
\cite{SanchezVega_phd}, a corpus specially built for evaluating the identification of paraphrase plagiarism. This corpus is an extension of the P4P corpus \cite{Barron2013CL}, which contains pairs of text fragments where one fragment represents the original source text and the other represents a paraphrased version of the original. In addition, the P4PIN corpus includes not paraphrase plagiarism cases, i.e., negative examples formed by pairs of unrelated texts samples with likely thematic or stylistic similarity. Table \ref{table:PlagiarimsExamples} shows two examples from this corpus, one case of paraphrase plagiarism and one of not-paraphrase plagiarism.

An important characteristic of this corpus is that each plagiarism case is labeled with a particular subtype of paraphrase. Authors of the P4P corpus \cite{Barron2013CL} employed a paraphrases typology, which includes four general classes, two of them with four sub-classes, for a total of nineteen types of paraphrases. For our purposes, we took two classes from the most general categorization level, and the four subclasses from the second categorization level as described below: 

\begin{itemize}
\item \textit{Morphology-based changes} include inflectional changes (\textit{e.g.,} affixes modification), modal verb modification (\textit{e.g.,} \textit{might $\rightarrow$ could}) and derivation changes.    

\item \textit{Lexicon-based changes} comprise modifications such as synthetic and analytic reconstruction, spelling and format change, polarity substitutions and converse substitutions; in general these types of changes alter only one lexical unit within a sentence preserving the original meaning.

\item \textit{Syntax-based modifications} cause structural alterations in a sentence, allowing to have the same meaning but redirecting the main focus to different elements within the sentence; paraphrase types included in this category are: diathesis alterations, negation switching, ellipsis, coordination changes and subordination with nesting changes.

\item \textit{Discourse-based modifications} alter the sentences' form and order; they include changes in punctuation marks, modifications in the syntactic structure, modality changes as well as some direct or indirect style alternations. 

\item \textit{Semantic-based changes} consider modifications involving substitution of some elements within a sentence that results in lexical and syntactical modifications without interfering with the original meaning of the sentence. Semantic-based changes represent the highest level of modifications. 

\item \textit{Miscellaneous-based changes} recollect all types of modifications that do not correspond to specific linguistic paraphrase phenomena, such as addition, deletion or changing the order of lexical units.
\end{itemize}

In summary, the P4PIN corpus has 2236 instances, where 75\% are not-plagiarism cases and 25\% are plagiarism cases.

\begin{table*}
\begin{center}
\caption{Examples of paraphrase-plagiarism and not-paraphrase-plagiarism in the P4PIN corpus. Underlined words represent common words between the original and the suspicious document; below each column appears the percentage of common words between text fragments.}
\label{table:PlagiarimsExamples}
\begin{scriptsize}
\begin{tabular}{|p{0.4in}|p{2.3in}|p{2.3in}|}
\hline
 & \textbf{Paraphrase plagiarism example} & \textbf{Not-paraphrase plagiarism example} \\
\hline
\textit{Original} & I pored through \uline{these pages, and as I} perused the lyrics \uline{of The Unknown Eros} that I had never read before, I appeared \uline{to have} found out something wonderful: there before \uline{me} \uline{was} an entire shining \uline{and} calming extract \uline{of} verses that were \uline{like} \uline{a new} universe \uline{to me}. &  \uline{The} fact \uline{that} \uline{an} omnipresent \uline{God} exists \uline{is} \uline{the} \uline{one universal} factor \uline{that} governs \uline{the laws of nature}. \uline{God} has set in place \uline{the laws} \uline{of} \uline{the} universe for His own purposes. \\
  & & \\
\textit{Suspicious}   &  \uline{I} dipped into \uline{these pages, and as I} \uline{read} for \uline{the} first time some \uline{of the} odes \uline{of The Unknown Eros}, \uline{I} seemed \uline{to have} made a great discovery: here \uline{was} \uline{a} whole glittering \uline{and} peaceful tract \uline{of} poetry which \uline{was} \uline{like} \uline{a} \uline{new} world \uline{to me}. &  \uline{The laws of nature are} \uline{the} art \uline{of} \uline{God}. Without \uline{the} presence \uline{of} such \uline{an} agent, \uline{one} who \uline{is} conscious \uline{of} all upon which \uline{the laws of nature} depend, producing all \uline{that} \uline{the laws} prescribe. \uline{The laws} themselves could have no existence. \\
 &  & \\
 \textit{Common words} &  $57.4\%$ &  $54.8\%$   \\
 \hline
\end{tabular}
\end{scriptsize}
\end{center}
\end{table*}

In order to get more insight on the relevance and robustness of the proposed measures we also evaluated them in the paraphrase identification task.\footnote{Although similar, paraphrase plagiarism identification differs from paraphrase identification in that the former is done with the intention of hiding the text-reuse (i.e., the plagiarism act)} For this purpose we used the well-known MSRP corpus \cite{dolan2005automatically}, which contains pairs of sentences labeled as ``mean the same thing” (paraphrase) or not (not-paraphrase) \cite{dolan2005automatically}. This corpus is divided in two partitions, a training set having 4,076 sentences pairs and a test set containing 1,725 examples; in both partitions, 67\% of the instances are plagiarism examples and the remaining 33\% are not-plagiarism cases. Contrary to the P4PIN, the MSRP corpus is not labeled by paraphrase sub-types.

\subsection{Semantic word similarity}
\label{sec:wordsimilarity}
Both proposed measures rely on the calculus of the semantic similarity or distance between pairs of words ($\text{sim}(x,y)$ or $\text{dist}(x,y)$). For the sake of simplicity we defined $\text{dist}(x,y) = 1 - \text{sim}(x,y)$. 

We used two different approaches for computing the word similarity. On the one hand, we used WordNet as knowledge source and applied the WUP similarity measure \cite{Wu94}. This measure calculates the semantic relatedness of two given words $x$ and $y$ by considering the depths of their synsets in the WordNet taxonomy ($s_x$ and $s_y$), along with the depth of their most specific common synset ($mcs$) as described by Formula \ref{eq:WUP}.

\begin{equation}
\label{eq:WUP}
\text{sim}(x, y) = \frac{2 * \text{depth}(\text{msc})}{\text{depth}(s_x) + \text{depth}(s_x)}
\end{equation}

On the other hand, we used the word2vec representation, and measured the similarity of words by means of the cosine function. In particular, we used the continuous Skip-gram model \cite{mikolov2013efficient} of the word2vec toolkit\footnote{\url{https://code.google.com/archive/p/word2vec/}} to generate the distributed representations of the words from the complete English Wikipedia. We considered 200-dimensional vectors, a context window of size 10, and 20 negative words for each sample.

\subsection{Classification process}
Once computed the similarity (or edit distance) between the suspicious and source texts, the next step is to determine whether or not the pair of texts are a case of plagiarism. When using the semantically-informed similarity measure, if the similarity score is greater than some threshold $\beta_s$, then the instance is classified as ``plagiarism'' otherwise the result is ``not-plagiarism''. On the other hand, when using the semantic-informed edit distance, if the distance score is greater that some threshold $\beta_d$, then the instance is labeled as ``not-plagiarism'' otherwise the result is ``plagiarism''.

For the experiments done with the P4PIN corpus we carried out a ten-fold cross-validation strategy. We considered as classification threshold ($\beta_s$ or $\beta_d$) the one that maximizes the classification performance at training. For the MSRP corpus we used the given training and test partitions. The classification threshold is defined from the training partition. In all the experiments we used the macro $F_1$-measure as main evaluation measure.

\section{Experimental Results}
\label{sec:Experiments}

This section presents the results of several experiments aimed to assess the effectiveness of the proposed measures in the task of paraphrase plagiarism identification, as well as to analyze their complementarity and their appropriateness for identifying plagiarism cases using different categories of paraphrases.

\subsection{Relevance of considering semantic information}
\label{subsec:semanticResults}

To assess the relevance of considering semantic information in the calculation of the similarity/distance between two texts, we carried out the following set of experiments: \textit{i}) using the original Jaccard coefficient (\textbf{J}),  ; \textit{ii}) using the original edit distance (\textbf{ED}); \textit{iii}) using the proposed semantically-informed measures with WordNet (\textbf{SJ-WN} and \textbf{SED-WN}) and with word2vec (\textbf{SJ-W2V} and \textbf{SED-W2V}).

\begin{table}[t]
\caption{$F_{1}$ results in the identification of paraphrase and paraphrase plagiarism,\\using the traditional and the proposed similarity and distance measures.\\Suffix W2V means word2vec and WN indicates WordNet.} 
\centering 
\begin{small}
\begin{tabular}{l@{~}c@{~}c@{~}c@{~}c@{~}c@{~}c@{~}} 
\toprule 
\textbf{Corpus} & \textbf{J} & \textbf{SJ-W2V} & \textbf{SJ-WN} & \textbf{ED} & \textbf{SED-W2V} & \textbf{SED-WN}\\
\midrule
P4PIN & 0.90& \textbf{0.91}&0.80 &0.87&0.90 & 0.82\\
MSRP & 0.80 & \textbf{0.81} &0.73& 0.75 &\textbf{0.81} & 0.76\\
\bottomrule
\end{tabular}
\end{small}
\label{tab:EXP1}
\end{table}

Results from Table \ref{tab:EXP1} show that the proposed semantically informed approaches, based on both the Jaccard and the Levenshtein edit distance measures, obtained better or equal $F_{1}$ results than the approaches using the original measures. This particularly happens when word2vec is used as word similarity function (SJ-W2V and SED-W2V). We attribute these results to the coverage of the semantic resources. Table \ref{tab:coverage} shows a comparative analysis of the vocabulary coverage for both WordNet and word2vec resources within each evaluated corpus. These results indicate that WordNet has lower coverage value than word2vec. Thus, results from Table \ref{tab:coverage} highlight the limitations of using an external resource such as WordNet.

\begin{table}[h]
\caption{Comparative analysis of the vocabulary coverage.}
\centering 
\begin{small}
\begin{tabular}{l@{~~~~~}c@{~~~~~}c@{~~~~~}}
\toprule 
\textbf{Corpus} & \textbf{WordNet} & \textbf{word2vec}\\
\midrule
P4PIN & 79.52\% & 91\% \\
MSRP & 79.1\% & 98\% \\
\bottomrule
\end{tabular}
\end{small}
\label{tab:coverage} 
\end{table}

\subsection{Complementary of the proposed measures}
\label{subsec:combinedmethods}

The proposed measures are similar in that both consider semantic information and, therefore, both can identify related texts even when they do not contain exactly matching words. However, they differ from each other in the way they compute the relatedness of texts. On the one hand, the similarity measure focuses on the \textit{content overlap}, whereas, on the other hand, the distance measure emphasizes the \textit{word order}. Accordingly, this section presents an experiment aimed to analyze the complementarity of the two measures.

\begin{figure*}
  \centering
  \includegraphics[width=170px]{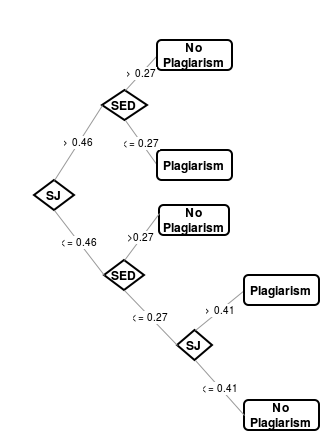}
  \caption{Decision tree of the combined approach on the P4PIN corpus.}
  \label{fig:P4PINTree}
\end{figure*}

The experiment reported in this section combines the best results from the previous section (i.e., SJ-W2V and SED-W2V). For the combination we used a supervised classification approach, where the scores obtained from both measures were used as features. We considered several learning algorithms, such as SVM, Naïve Bayes and J48, but we only report the results obtained by J48 because they outperformed the others as well as allow us to understand the classification criteria (refer to Figure~\ref{fig:P4PINTree}).

Table \ref{tab:Expsj48} shows the results from this experiment. It can be noticed that the results obtained by the combined approach clearly outperform the results from the approaches using the proposed measures individually. Hence, our preliminary conclusion is that these two measures are in fact complementary to each other. Additionally, this table shows the state-of-the-art results for the two used datasets. As noticed, the results from our combined approach are close to the reference results, nonetheless, ours is a much more simple approach (for example, \cite{cheng2015syntax} reports a recursive neural network using syntax-aware and multi-sense word embeddings).  

\begin{table}
\caption{$F_{1}$ results from the combination of the semantically-informed similarity and distance measures. The SOA column indicates the state-of-the-art performance reported for each dataset.} 
\centering
\begin{small}
\begin{tabular}{l@{~~}c@{~~}c@{~~}c@{~~}c@{~~}} 
\toprule
\textbf{Corpus} & \textbf{SJ-W2V} & \textbf{SED-W2V} & \textbf{\textit{Combined}} & \textbf{SOA}
\\
\midrule
P4PIN & 0.90& 0.91 & \textbf{0.93} & 0.92 \cite{SanchezVega_phd}\\
MSRP & 0.81 & 0.81& \textbf{0.83} & 0.85 \cite{cheng2015syntax}\\
\bottomrule
\end{tabular}
\end{small}
\label{tab:Expsj48} 
\end{table}

\subsection{Robustness on different paraphrase categories}
\label{subsec:paraphraseTypes}

The plagiarism examples from the P4PIN corpus are categorized according to their paraphrases types, namely: \textit{morphology, lexicon, syntax, discourse, semantic} and \textit{miscellaneous} changes \cite{Barron2013CL} (refer to Section \ref{subsec:dataset}). The experiments reported in this section aim at measuring the robustness of the proposed semantically-informed measures against different paraphrase practices. Table \ref{tab:EXP3} shows the obtained results.

These results indicate that the proposed measures (using word2vec as semantic resource) consistently improve the performance results of the traditional variants. They also indicate that paraphrases from the \textit{semantic} category are the harder to identify. This performance was expected, since semantic changes involve lexical and syntactical modifications. Additionally, these results outperform the state-of-the-art  in all categories, evidencing that the supervised combined approach is the best option for identifying plagiarism regardless of the type of paraphrase. 

\subsection{On the complexity of corpora}
\label{subsec:dificulty}

In order to provide a deeper analysis on the obtained results, we decided to investigate the level of complexity of the employed corpora. Through this analysis we aim to figure out under which circumstances our proposed semantically informed metrics perform the better. 

\begin{table*}[t]
\caption{$F_{1}$ results in several paraphrase categories using different similarity and distance measures. The SOA column shows state-of-the-art results reported in \cite{SanchezVega_phd}. In  \cite{SanchezVega_phd} character n-grams are used for representing the documents and measuring their similarity.}
\centering 
\begin{tabular}{l@{~~~~}c@{~~~~}c@{~~~~}c@{~~~~}c@{~~~~}c@{~~~~}c@{~~~~}} 
\toprule 
\multirow{2}{1in}{\textbf{Paraphrases categories}}&\multicolumn{2}{c}{\textit{Jaccard}}&\multicolumn{2}{c}{\textit{Levenshtein}}& \multirow{2}{*}{\textbf{\textit{Combined}}}& \textbf{SOA}\\
\cmidrule(r){2-3}
\cmidrule(r){4-5}
& \textbf{J} & \textbf{SJ-W2V}&\textbf{ED} & \textbf{SED-W2V} & &\cite{SanchezVega_phd}\\
\midrule
Morphological & 0.85  & 0.88 & 0.85 & 0.86 & \textbf{0.92} & 0.90\\
Lexical       & 0.90  & 0.91 & 0.88 & 0.89 & \textbf{0.93} & 0.92\\
Syntactical   & 0.88  & 0.89 & 0.85 & 0.87 & \textbf{0.93} & 0.91\\
Discourse     & 0.86  & 0.87 & 0.86 & 0.89 & \textbf{0.92} & 0.89\\		 
Semantic      & 0.77  & 0.78 & 0.73 & 0.80 & \textbf{0.83} & 0.77\\ 
Miscellaneous & 0.89  & 0.89 & 0.85 & 0.87 & \textbf{0.92} & 0.90\\
\bottomrule
\end{tabular}
\label{tab:EXP3} 
\end{table*}

For determining the level of complexity of a given corpus $C$ we propose the following straightforward measure (refer to Formula \ref{eq:corpusComplex}), which assesses the lexical concordance ($\text{LC}$) across both plagiarism and not-plagiarism examples. 

\begin{equation}
\label{eq:corpusComplex}
\text{LC}(C)=\frac{|C_{\text{neg}}|-\text{O}(C_{\text{neg}})+\text{O}(C_{\text{pos}})}{|C|}
\end{equation}

where $C_{\text{neg}}$ and $C_{\text{pos}}$ represent the negative and positive partitions of corpus $C$ respectively. Accordingly, \text{O}$(C_{x})$ represents the accumulated similarity between all pairs of documents contained in the $x$ partition of the corpus $C$ and it is obtained using the Formula \ref{eq_overlap}, where \text{J}$(A,B)$ represents the Jaccard coefficient between the pair of documents $A$ and $B$. 

\begin{equation}
\label{eq_overlap}
\text{O}(C_{x})= \sum_{\forall(A,B) \in C_{x}} \text{J}(A,B)
\end{equation}

The closer the value of lexical concordance to zero means the corpus is more complex, whilst the closer to one indicated an easier corpus. For example, in a low complexity corpus ($\text{LC}(C) \to 1$) the positive instances are merely verbatim cases and the negative examples are completely unrelated text chunks.

Table \ref{tab:REP} shows the $\text{LC}$ values for the MSRP and P4PIN collections. It can be noticed that MSPR is more complex than P4PIN (see first two rows from Table \ref{tab:REP}). Additionally, in the P4PIN corpus we observe that the more complex paraphrase category is the semantic category, whereas the easier is the lexical one.

As a final experiment we analyze the influence of the complexity of the collections over the performance of the proposed semantic enriched measures. In particular we analyzed the correlation between the $\text{LC}$ value of each category of the P4PIN corpus and the $F_1$ improvement of the proposed approach over the baselines. For this analysis we applied the Spearman Correlation Coefficient.

\begin{table}[h]
\caption{Lexical concordance values of the employed corpora} 
\centering 
\begin{tabular}{l@{~~~~~}c@{~~~~~}} 
\toprule
\textbf{Corpus} & \textbf{LC value}\\
\hline 
P4PIN & 0.76\\
MSRP & 0.56\\
\midrule
\textbf{Paraphrase types} & \textbf{LC value}\\
\midrule
Lexical & 0.41\\
Discourse & 0.41\\
Miscellaneous & 0.39\\
Syntactical & 0.39\\
Morphological & 0.38\\
Semantic & 0.29\\ 
\bottomrule
\end{tabular}
\label{tab:REP}
\end{table}

Table \ref{tab:correlation} shows the obtained correlation results, indicating some very interesting insights from the proposed measures.
On the one hand, there is a strong correlation between the complexity of the corpus and the performance of our combined method. Given the correlation is negative, it indicates that the more complex is the corpus (the smallest the $\text{LC}$ value), the greater is the advantage of our method over SOA results; in other words, our proposed method performs consistently better when the corpus has a high complexity level. A similar situation occurs when employing our semantically informed edit distance (SED) approach; it especially outperforms the ED results for the complex paraphrase categories. On the other hand, the correlation results indicate that the improvement of SJ-W2V over J is not related to the corpus complexity. 

\begin{table}[h!]
\caption{Correlation analysis} 
\centering 
\begin{tabular}{l@{~~~~~}c@{~~~~~}} 
\toprule
\textbf{Compared methods} & \textbf{$r$}\\
\hline 
SJ-W2V \textit{vs.} J & -0.0377\\
SED-W2V \textit{vs.} ED & -0.8771\\
\textit{Combined} \textit{vs.} SOA & -0.8985\\
\bottomrule
\end{tabular}
\label{tab:correlation}
\end{table}

\section{Conclusions and future work}
\label{sec:Conclusions}

We have introduced an approach for paraphrase plagiarism detection which proposes the inclusion of semantic information to traditional similarity and edit distance measures. The aim of the proposed semantically-informed measures is to allow assessing the relatedness between suspicious and source texts even when they do not contain exactly matching words.

We hypothesized that using the proposed semantically-informed measures, a method for paraphrase plagiarism identification would be more accurate in solving the task. Performed experiments indicate that our proposed method obtained state-of-the-art results, especially when distributed word representations are considered as a semantic resource. Additionally, experiments demonstrated that the information provided by the two semantically-informed measures is complementary to each other, resulting in useful features for a supervised classifier to learn whether or not the pair of texts are a case of plagiarism. Further, we investigated the degree of robustness of the proposed measures against different subtypes of paraphrase plagiarism. Obtained results showed that the proposed approaches, either individually or combined, are able to improve the performance of traditional techniques for the distinct paraphrase plagiarism categories, particularly for those with higher complexities. Finally, it is important to highlight that obtained results are competitive to those reported in recent research works, but, in contrast, the proposed approach represents a much more simple method.

As future work we plan to study the sensitivity of our method to the coverage of the semantic resource, in particular we plan to evaluate our method using a word2vec representation trained over a larger corpus.\\

\noindent\textbf{Acknowledgements:} This work was partially supported by CONACYT under scholarship 401887, project grants 257383, 258588 and 2016‐01‐2410 and under the Thematic Networks program (Language Technologies Thematic Network project 281795). The work of the fourth author was partially supported by the SomEMBED TIN2015-71147-C2-1-P MINECO research project and by the Generalitat Valenciana under the grant ALMAMATER (Prometeo II/2014/030). 

\bibliographystyle{plain}
\bibliography{references}
\end{document}